\documentclass{article}
\usepackage{spconf,amsmath,graphicx}

\usepackage{epsfig}
\usepackage{amssymb}
\usepackage[linesnumbered,vlined,ruled]{algorithm2e}

\title{CLUSTERING IMAGES BY UNMASKING -- A NEW BASELINE}
\name{Mariana-Iuliana Georgescu \qquad {Radu Tudor} Ionescu}
\address{Faculty of Mathematics and Computer Science, University of Bucharest, Romania}
\begin{document}
%
\maketitle
\begin{abstract}
We propose a novel agglomerative clustering method based on \emph{unmasking}, a technique that was previously used for authorship verification of text documents and for abnormal event detection in videos. In order to join two clusters, we alternate between $(i)$ training a binary classifier to distinguish between the samples from one cluster and the samples from the other cluster, and $(ii)$ removing at each step the most discriminant features. The faster-decreasing accuracy rates of the intermediately-obtained classifiers indicate that the two clusters should be joined. To the best of our knowledge, this is the first work to apply unmasking in order to cluster images. We compare our method with k-means as well as a recent state-of-the-art clustering method. The empirical results indicate that our approach is able to improve performance for various (deep and shallow) feature representations and different tasks, such as handwritten digit recognition, texture classification and fine-grained object recognition.
\end{abstract}
\section{Introduction}
\label{sec:intro}

\emph{Unmasking} \cite{Koppel-JMLR-2007} is an unsupervised method which is based on testing the degradation rate of the cross-validation accuracy of learned models, as the best features are iteratively dropped from the learning process. Koppel et al. \cite{Koppel-JMLR-2007} offered evidence that this unsupervised technique can solve the authorship verification problem with very high accuracy. More recently, Ionescu et al. \cite{Ionescu-ICCV-2017} have used unmasking to detect abnormal events in video, without requiring any training data. To the best of our knowledge, we are the first to propose a clustering approach based on unmasking. Our approach falls in the category of agglomerative clustering methods, and it joins two clusters by alternating between two steps: $(i)$ training a binary classifier to distinguish between the samples from one cluster and the samples from the other cluster, and $(ii)$ removing at each step the most discriminant features. If the samples from the two clusters belong to the same class, than the samples become indistinguishable when discriminant features are gradually removed. Hence, the faster-decreasing accuracy rates of the intermediately-obtained classifiers indicate that the two clusters should be joined.

We conduct experiments on the MNIST~\cite{lecun-bottou-ieee-1998}, the UIUCTex~\cite{Lazebnik-PAMI-2005} and the Oxford Flowers~\cite{Nilsback-CVPR-2006} data sets in order to compare the proposed clustering method with the k-means clustering algorithm. On the MNIST data set, we also consider as baseline a recent method presented in~\cite{Guo-IJCAI-2017}. The empirical results indicate that our approach can significantly outperform all these baselines.

The paper is organized as follows. Related works are presented in Section~\ref{sec:relatedart}. Our approach is presented in Section~\ref{sec:method}. The experiments are discussed in Section~\ref{sec:experiments}. We draw our conclusions in Section~\ref{sec:conclusion}.

\vspace{-0.25cm}
\section{Related work}
\label{sec:relatedart}
\vspace{-0.15cm}

Researchers have made considerable effort~\cite{Guo-IJCAI-2017,zhang-acm-1996,huang-1998,McCallum-2000,bohm-2004,kailing-2004,achtert-2006,kriegel-2009,Yang-TIP-2010,Han-DM-2011,dinu-radu-iconip-cluster-2012,radu-NCA-2013,Faktor-TPAMI-2014,Tian-AAAI-2014,Xie-ICML-2016,Cao-TNNLS-2017,Yang-ICML-2017,Li-TNNLS-2018} to improve the performance over the commonly-used clustering algorithms, such as k-means. Hence, various clustering algorithms have emerged that differ significantly in how they form the data clusters. Clustering methods can be roughly divided into several categories, including hierarchical clustering methods~\cite{zhang-acm-1996,achtert-2006}, partitioning methods~\cite{huang-1998,dinu-radu-iconip-cluster-2012,radu-NCA-2013,Cao-TNNLS-2017,Li-TNNLS-2018,Ng-NIPS-2002}, density-based methods~\cite{bohm-2004,kailing-2004}, and deep learning methods~\cite{Guo-IJCAI-2017,Tian-AAAI-2014,Xie-ICML-2016,Yang-ICML-2017}. Some researchers have focused on addressing specific clustering problems, including clustering large data sets~\cite{zhang-acm-1996,McCallum-2000}, clustering data with categorical values~\cite{huang-1998,Cao-TNNLS-2017}, subspace clustering~\cite{kailing-2004} and correlation clustering~\cite{bohm-2004,kriegel-2009}.
With the recent need to process larger and larger data sets, the willingness to trade the semantic meaning of the generated clusters for computational performance has been increasing. This led to the development of pre-clustering methods such as Canopy clustering~\cite{McCallum-2000}, which can process huge data sets efficiently, but the resulting clusters are only a rough pre-partitioning of the data set. These partitions can be subsequently analyzed with existing slower methods such as k-means clustering. For high-dimensional data, many of traditional clustering methods fail due to the curse of dimensionality, which renders particular distance functions problematic in high-dimensional spaces. This led to new clustering algorithms for high-dimensional data that focus on subspace clustering~\cite{kailing-2004} and correlation clustering~\cite{kriegel-2009}. 
The reader is referred to~\cite{Han-DM-2011} for a complete review of the major clustering methods.

More closely-related to our work, are recent methods focused particularly on clustering images~\cite{Guo-IJCAI-2017,Yang-TIP-2010,Faktor-TPAMI-2014,Xie-ICML-2016,Yang-ICML-2017}. While some researchers have focused strictly on the clustering task~\cite{Yang-TIP-2010,Faktor-TPAMI-2014}, others considered learning unsupervised image embeddings using neural networks~\cite{Yang-ICML-2017,Caron-ECCV-2018} or auto-encoders~\cite{Guo-IJCAI-2017,Xie-ICML-2016,Ji-NIPS-2017}. Different from the recent approaches focused on learning deep image embeddings, our work is particularly focused on the clustering task. As shown in the experiments, various features can be plugged into our clustering framework, including features that are trained in an unsupervised manner~\cite{Caron-ECCV-2018}.

\vspace{-0.25cm}
\section{Clustering by unmasking}
\label{sec:method}
\vspace{-0.15cm}

\emph{Clustering} is the task of assigning a set of objects into groups (termed clusters) such that the objects in the same cluster are more similar to each other than to those in other clusters. We approach the clustering task by proposing an agglomerative clustering algorithm that joins clusters based on unmasking. Our algorithm takes as input a set of $m$ training samples and the number of desired clusters $k$, and provides as output the cluster assignments of the given samples, as well as the resulted cluster centroids, that can be used to cluster new (test) samples based on the distance to the closest centroid. Our algorithm also takes as input some additional parameters required by the unmasking technique, which will be explained next. Unmasking is based on estimating how well a linear classifier can distinguish between the samples of two clusters, which means that the clusters must contain at least a few data samples to begin with. Therefore, our algorithm starts with an initial number of clusters $K$, with $K \leq m$ and $K >> k$. In the experiments, we tried two alternative approaches to form the initial clusters. One approach is to randomly select $K$ data points as cluster centroids, and assign the rest of the data points to clusters based on the Euclidean distance to the nearest centroid. This approach is equivalent to the standard k-means initialization of the clusters, the only difference being that $K >> k$. The other approach is based on setting $K = m$ and on generating artificial data samples from the original data samples by adding various transformations such as Gaussian noise, horizontal or vertical flips, rotations, random crops and illumination changes. Each artificially-generated sample is assigned to the same cluster as the original image it came from. We note that the second approach is particularly suitable for clustering image samples, but it constrains our algorithm to be applied only on image data.

After building the initial $K$ clusters, we start joining the clusters until we end up with the desired number of clusters $k$. For each pair of clusters $i$ and $j$, we compute a score that indicates the likelihood of the statement \emph{``clusters $i$ and $j$ should be joined''} to be true. In order to compute this score, we apply unmasking, as described next. We initially assume that the samples in cluster $i$ belong to a different class than the samples in cluster $j$, and we assign binary labels accordingly. The goal of applying unmasking is to test whether this assumption (hypothesis) is actually true, by iteratively training and testing a binary classifier to distinguish between the labeled training samples, while removing the most discriminant features at each iteration. Before applying the classifier, the samples in each cluster are randomly split into a training set and a testing set of equal size. Hence, the training set contains half of the positively-labeled samples from cluster $i$ and half of the negatively-labeled samples from cluster $j$, while the testing set contains the rest of the data samples from clusters $i$ and $j$. Next, we $(i)$ train a linear Support Vector Machines (SVM) classifier on the training set (until convergence) and evaluate it on the testing set, retaining the accuracy rate. We note that although SVM is a supervised approach, it is never trained on ground-truth labels. We then $(ii)$ sort the weights of the SVM by their absolute values, in descending order, we take the first $s$ indexes of the sorted list, and we remove the corresponding features from all the samples in the training and testing sets. Since the classifier assigned higher weights (in absolute value) to these features, it means that the removed features are the most discriminant. We repeat steps $(i)$ and $(ii)$ for $n$ iterations, retaining the accuracy rate at each iteration. Since we remove features at each iteration, the accuracy rate of the binary classifier naturally tends to drop over time. However, if the samples in the two clusters belong to different classes, i.e. the assumed hypothesis is true, then the accuracy rate will drop at a slower pace. On the other hand, if the samples in the two clusters belong to the same class,  i.e. the assumed hypothesis is false, then the classifier will have a hard time in distinguishing between samples as features get removed, and the accuracy rate will drop faster. Hence, the score of joining clusters $i$ and $j$ is computed as one minus the accuracy rate averaged over the $n$ iterations. Along with $K$, the number of unmasking iterations $n$ and the number of discriminant features $s$ that are removed at each iteration represent input parameters of our algorithm.

After computing the score for each pair of clusters, we merge each cluster $i$ with a cluster $j$ (using a Greedy approach), such that the score of joining clusters $i$ and $j$ is maximum, for all $j \in \{1, 2, ...., K\}$, with $j \neq i$. If we reach the number of clusters $k$ at any point during the merging process, we halt the execution and return the current cluster assignments. Otherwise, we continue by computing the merging scores for the newly-formed clusters. Our algorithm ends when the current number of clusters is equal to $k$. Along with the cluster assignments, our algorithm also returns the centroid for each cluster. In the testing stage, a new sample (not seen during clustering) is assigned to a cluster, if the Euclidean distance to the respective cluster centroid is minimum. 

\vspace{-0.25cm}
\section{Experiments}
\label{sec:experiments}
\vspace{-0.15cm}

\noindent
{\bf Data sets.}
The first data set used in the evaluation is MNIST~\cite{lecun-bottou-ieee-1998}. The MNIST database contains $60,000$ train samples and $10,000$ test samples, size-normalized to $20 \times 20$ pixels, and centered by center of mass in $28 \times 28$ fields.

The second set of experiments are conducted on the UIUCTex data set~\cite{Lazebnik-PAMI-2005}. UIUCTex contains $1000$ texture images of $640 \times 480$ pixels representing different types of textures such as bark, wood, floor, water, and more. There are $25$ classes, with $40$ texture images per class. Textures are viewed under significant scale, viewpoint and illumination changes. Images also include non-rigid deformations. The data set is split into $50\%$ for training and $50\%$ for testing. 

The third data set included in the evaluation is the Oxford Flowers data set~\cite{Nilsback-CVPR-2006} with $17$ categories. 
There are a total of $1360$ images in the data set, with a number of $80$ images per category. The data set is split into $50\%$ for training and $50\%$ for testing.

\noindent
{\bf Baselines.}
In all experiments, we consider k-means as a baseline clustering method. We also consider as baseline a stub version of our clustering by unmasking approach, which performs only $n=1$ iterations, in order to demonstrate the utility of training the binary SVM and removing the discriminant features into several iterations, according to the full version of our algorithm. On the MNIST data set, there are some unsupervised methods~\cite{Guo-IJCAI-2017,Xie-ICML-2016,Yang-ICML-2017} that report results, but only on the entire set, which includes both training and testing. However, to demonstrate the generalization capacity of our clustering algorithm, we use the official training and testing split. In this context, we choose as baseline one recent work~\cite{Guo-IJCAI-2017} that provides the code online at {https://github.com/XifengGuo/IDEC}. We used the provided code in order to compute the accuracy rate of the Improved Deep Embedded Clustering (IDEC)~\cite{Guo-IJCAI-2017} on the MNIST test set. We did not find any results reported by recent clustering methods on UIUCTex or Oxford Flowers, perhaps because small data sets are generally more challenging for unsupervised methods, as also noted by Faktor et al.~\cite{Faktor-TPAMI-2014}. In each and every experiment, we include the results of the random chance baseline as well as the results of the supervised SVM.

\noindent
{\bf Features.}
Since our algorithm is designed strictly for clustering (not for unsupervised learning of feature embeddings), we can plug in various deep or handcrafted features. For the MNIST data set, we consider the raw pixels as features. However, for the UIUCTex and the Oxford Flowers data sets containing natural images, raw pixels are not discriminative enough. We thus consider three different kinds of features, as described next. 
First of all, we consider deep supervised features provided by the first fully-connected layer of VGG-f~\cite{Chatfield-BMVC-14}, known as \emph{fc6}, resulting in a $4096$-dimensional feature vector for each image. Although VGG-f is pre-trained on a different task~\cite{Russakovsky2015}, the clustering method can no longer be considered a fully-unsupervised approach when VGG-f features are plugged in. Nevertheless, the method can be viewed as an \emph{unsupervised transfer learning approach}. We remove the \emph{daisy} and the \emph{iris} classes from the Oxford Flowers data set, since these classes are already learned by the VGG-f model.
Second of all, we consider handcrafted features computed with a standard bag-of-visual-words (BOVW) model~\cite{Ionescu_Popescu_PRL_2014} based on dense SIFT descriptors~\cite{Lowe-SIFT-2004,bosch-phow-2007}. The descriptors are quantized into visual words by k-means clustering. The visual words are then stored in a randomized forest of k-d trees to reduce search cost. Finally, a histogram of visual words is computed for each image, by counting the occurrences of each visual word in the image. 
In our experiments, we use $4000$ visual words. In addition, for the Oxford Flowers data set, we include color descriptors along with the SIFT descriptors.
Third of all, we consider deep unsupervised features that are extracted from the pre-trained AlexNet architecture provided by Caron et al.~\cite{Caron-ECCV-2018}. In their experiments, Caron et al.~\cite{Caron-ECCV-2018} showed that the activation maps produced by the mid-level convolutional layers yield optimal accuracy rates. Therefore, we extract features from the \emph{conv3} layer of their unsupervised neural network. The final features are obtained by adding a $4 \times 4$ max-pooling layer after the \emph{conv3} layer, which produces a $3456$-dimensional feature vector for each image. 
Finally, we would like to stress out that the handcrafted BOVW features and the deep features provided by the pre-trained model of Caron et al.~\cite{Caron-ECCV-2018} are obtained in a completely unsupervised fashion.

\noindent
{\bf Parameter tuning.}
We tune all parameters on the training sets. On the MNIST data set, we choose $K=1000$, while for the other data sets, we set $K$ equal to the number of training samples. For the number of unmasking iterations, we experiment with $n \in \{4,5,6,7,8,9,10\}$. In each case, the number of features to be removed is set depending on $n$ and on the dimension of the feature vectors, such that in the last iteration, there are still some features left for training the SVM.

\noindent
{\bf Evaluation metrics.}
We use the same evaluation protocol as in previous works~\cite{Guo-IJCAI-2017,Yang-TIP-2010,Xie-ICML-2016,Yang-ICML-2017} and employ the unsupervised clustering accuracy (ACC) and the Normalized Mutual Information (NMI) score as performance metrics. For the ACC metric, the optimal mapping between the ground-truth assignments and the cluster assignments provided by an unsupervised algorithm needs to determined using the Hungarian algorithm. We report results only on the test sets.

\begin{table}[!t]
\small{
\caption{Clustering performance of various baselines versus clustering by unmasking on the MNIST test set. Higher ACC or NMI values are better.}\label{tab_results_mnist}
\vspace{0cm}
\begin{center}
\begin{tabular}{|l|c|c|c|c|}
\hline
Method                          & ACC           & NMI   \\    
  
\hline
\hline
Random chance                   & $10.00\%$     & - \\
SVM                             & $94.40\%$     & - \\    
\hline 
K-means                         & $55.82\%$     & $52.18\%$ \\   
IDEC~\cite{Guo-IJCAI-2017}      & $71.45\%$     & $69.40\%$ \\  
\hline 
Unmasking (n=1)                 & $72.58\%$     & $64.99\%$ \\     
Unmasking                       & $81.40\%$     & $69.76\%$ \\
\hline
\end{tabular}
\end{center}
}
\vspace{-0.8cm}
\end{table}

\begin{figure}[!t]

\begin{center}
\includegraphics[width=0.95\linewidth]{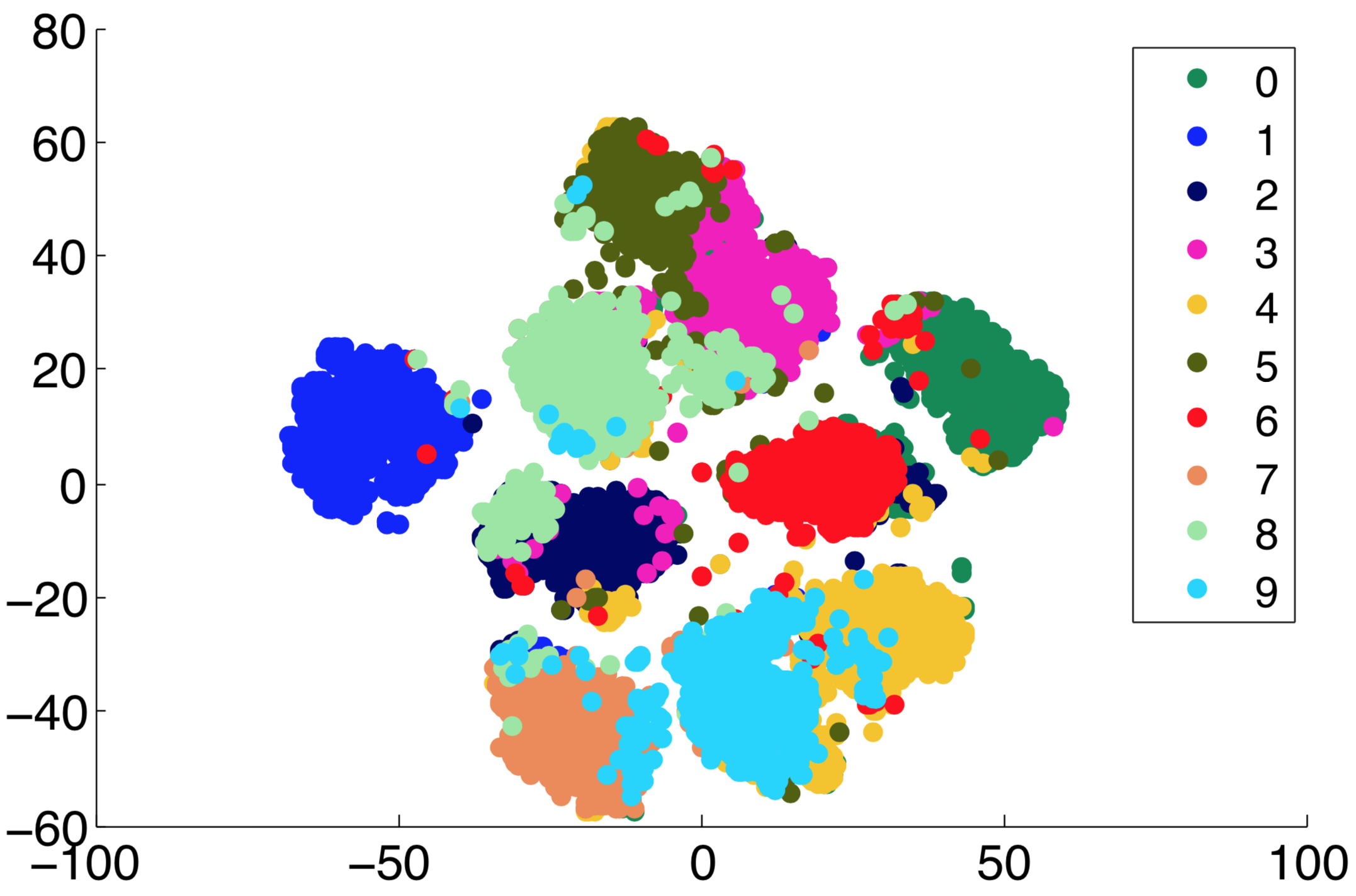}
\end{center}
\vspace*{-0.7cm}
\caption{Visualization of unmasking-based clustering results on the MNIST test set. Best viewed in color.}
\label{fig_tsne}
\vspace*{-0.3cm}
\end{figure}

\begin{table}[!t]
\setlength\tabcolsep{2.5pt}
\small{
\caption{Clustering performance of various baselines versus clustering by unmasking on the UIUCTex and the Oxford Flowers test sets. Higher ACC or NMI values are better.}\label{tab_results}
\vspace{0cm}
\begin{center}
\begin{tabular}{|l|l|c|c|c|c|}
\hline 
Features &  Method          & \multicolumn{2}{|c|}{UIUCTex} & \multicolumn{2}{|c|}{Oxford Flowers} \\
\cline{3-6}  &              & ACC           & NMI           & ACC           & NMI \\
\hline
\hline
-       & Random chance     & $4.00\%$      & -             & $6.67\%$      & - \\
\hline
        & SVM               & $97.20\%$     & -             & $95.50\%$     & - \\ 
\cline{2-6}
VGG-f   & K-means           & $48.20\%$     & $70.15\%$     & $60.35\%$     & $69.55\%$ \\
        & Unmasking (n=1)   & $19.80\%$     & $54.81\%$     & $45.50\%$     & $62.98\%$ \\ 
        & Unmasking         & $61.40\%$     & $74.94\%$     & $67.50\%$     & $75.82\%$ \\ 
\hline
        & SVM               & $94.60\%$     & -             & $80.83\%$     & - \\ 
\cline{2-6}
BOVW    & K-means           & $25.40\%$     & $46.83\%$     & $22.10\%$     & $22.83\%$ \\
        & Unmasking (n=1)   & $35.20\%$     & $55.30\%$     & $12.83\%$     & $14.64\%$ \\ 
        & Unmasking         & $44.60\%$     & $58.81\%$     & $25.00\%$     & $25.37\%$ \\ 
\hline
        & SVM               & $96.20\%$     & -             & $81.00\%$     & - \\
\cline{2-6}
AlexNet & K-means           & $36.80\%$     & $58.52\%$     & $26.89\%$     & $30.43\%$ \\
        & Unmasking (n=1)   & $34.20\%$     & $62.07\%$     & $9.80\%$      & $18.19\%$ \\ 
        & Unmasking         & $48.60\%$     & $69.78\%$     & $33.33\%$     & $38.00\%$ \\ 
\hline
\end{tabular}
\end{center}
}
\vspace{-0.8cm}
\end{table}

\noindent
{\bf Results.}
Table~\ref{tab_results_mnist} shows the results obtained on the MNIST test set. In terms of accuracy, our approach surpasses k-means by more than $25\%$ and the state-of-the-art IDEC approach~\cite{Guo-IJCAI-2017} by nearly $10\%$. We note that the performance improvements over k-means and IDEC are statistically significant, according to a paired McNemar's test performed at a significance level of $0.01$. Using a single iteration for unmasking ($n=1$) yields results that are comparable to those obtained by IDEC. This shows the importance of applying unmasking over multiple iterations, in this particular case $n=8$. We note that the performance gap between the supervised SVM and our approach is greater than $10\%$. In terms of NMI, our clustering method outperforms all unsupervised baselines. We provide a t-SNE~\cite{Maaten-JMLR-2008} visualization of our MNIST clustering results in Figure~\ref{fig_tsne}. The figure shows that our algorithm provides well-formed clusters for digits $0$, $1$, $2$, $5$, $6$ and $7$, while making some confusions between the pair of clusters $3$ and $8$, and the pair of clusters $4$ and $9$.

Table~\ref{tab_results} presents the results obtained on the UIUCTex and the Oxford Flower test sets. The results are generally higher in the unsupervised transfer learning setting, i.e. when we use deep features from the supervised VGG-f model. In the fully-unsupervised setting, i.e. when we use BOVW or AlexNet features, all methods obtain slightly better results with the unsupervised AlexNet features. Nonetheless, our clustering by unmasking approach obtains better performance than k-means, irrespective of the feature set. We note that our approach attains the highest improvements over k-means on the UIUCTex data set. For instance, using the BOVW features, the improvement in terms of accuracy  over k-means is slightly higher than $19\%$. Nevertheless, our accuracy is significantly higher than the accuracy of k-means on both data sets, according to a paired McNemar's test performed at a significance level of $0.01$. The baseline that uses a single iteration for unmasking ($n=1$) yields lower performance, even compared to k-means, in all except one case. One again, these results show the importance of applying unmasking over multiple iterations, as designed by Koppel et al. \cite{Koppel-JMLR-2007}.

\vspace{-0.25cm}
\section{Conclusion}
\label{sec:conclusion}
\vspace{-0.15cm}

In this paper, we presented an agglomerative clustering approach based on unmasking. We compared our approach with k-means as well as a recent unsupervised method~\cite{Guo-IJCAI-2017}, showing significantly better results in image clustering, using various deep and handcrafted features. In future work, we aim to evaluate our clustering method on other kinds of data, and to use other binary classifiers (instead of SVM) for unmasking.
\vspace{-0.15cm}


\bibliographystyle{IEEEbib}
\bibliography{references}

\end{document}